\pdfoutput=1

\documentclass[11pt]{article}

\usepackage{acl}

\usepackage{times}
\usepackage{latexsym}

\usepackage[T1]{fontenc}

\usepackage[utf8]{inputenc}

\usepackage{microtype}
\usepackage[flushleft]{threeparttable}
\usepackage{booktabs}
\usepackage{xspace}
\usepackage{arydshln}
\usepackage{graphics}
\usepackage{graphicx}
\usepackage{amsmath}
\usepackage{enumitem}
\usepackage{caption}
\usepackage{subcaption}
\usepackage{multirow}
\usepackage{color}
\usepackage{paralist}
\usepackage[ruled,vlined]{algorithm2e}
\usepackage{soul}

\usepackage{lipsum}
\DeclareMathOperator*{\argmax}{arg\,max}

\definecolor{darkgreen}{rgb}{0.0, 0.4, 0.13}

\newcommand{\sys}{\textsc{Pinocchio}\xspace}
\newcommand{\secref}[1]{\S\ref{#1}}
\newcommand{\githuburl}[1]{\url{https://github.com/allenai/pinocchio}}

\title{Don't Say What You Don't Know: Improving the Consistency of Abstractive Summarization by Constraining Beam Search}

\author{Daniel King\thanks{\ \ Both authors contributed equally. The first author's work was performed while at the Allen Institute for AI.}\textsuperscript{$*^{\dagger}$} \quad 
        Zejiang Shen\textsuperscript{$*^{\ddagger}$}\quad 
        Nishant Subramani$^{\star,\diamondsuit}$ \quad 
        Daniel S. Weld$^{\star,\heartsuit}$ \quad \\
        {\bf
        Iz Beltagy$^\star$ \qquad
        Doug Downey$^{\star,\clubsuit}$ \quad 
        }
        \\ [1mm]
        $^\dagger$MosaicML \quad
        $^\ddagger$MIT \quad
        $^\star$Allen Institute for AI \quad $^\diamondsuit$Masakhane \\
        \quad $^\heartsuit$University of Washington \quad $^\clubsuit$Northwestern University
        \\
        {\tt\small daking@hey.com \ zjshen@mit.edu \ \{nishants, danw, beltagy, dougd\}@allenai.org}
}

\begin{document}
\maketitle
\begin{abstract}

Abstractive summarization systems today produce fluent and relevant output, but often ``hallucinate'' statements not supported by the source text. %
We analyze the connection between %
hallucinations and training data, and find evidence that 
models hallucinate because they train on target summaries that are unsupported by the source.
Based on our findings, we present \sys, a new decoding method that improves the consistency of a transformer-based abstractive summarizer by constraining beam search to avoid hallucinations. 
Given the model states and outputs at a given step, \sys\  detects likely model hallucinations
based on %
various measures of attribution to the source text. 
\sys\ backtracks to find more consistent output,
and can opt to produce no summary at all 
when %
no consistent generation can be found. 
In experiments, we find that \sys\ improves the consistency of generation by an average of~68\% on two abstractive summarization datasets, without hurting recall.
\end{abstract}

\section{Introduction}

Abstractive text generation is an important task with the promise of compressing lengthy source material into concise summaries, satisfying application or user needs.  
Pretrained abstractive summarizers (e.g. BART \citep{lewis-etal-2020-bart}) have recently achieved new state-of-the-art (SOTA) across multiple datasets \citep{fabbri2020summeval}. 
However, these systems remain unusable in most real world scenarios, because they frequently hallucinate information that is inconsistent with the input~\citep{maynez-etal-2020-faithfulness}.

Many researchers have proposed methods to assess and improve the consistency\footnote{We use the terms ``consistent'' and "hallucinated" as antonyms, and avoid ``factual''. Check Section~\ref{sec:related} for details.
} 
of summarization  systems. 
Two popular approaches are 1) incorporating extracted knowledge \citep{Zhu2020EnhancingFC} (possibly in the form of questions \citep{durmus-etal-2020-feqa}), and 2) incorporating a consistency text classifier \citep{kryscinski-etal-2020-evaluating} (often based on natural language inference (NLI) \citep{falke-etal-2019-ranking}). 
These methods tend to reduce the problem of generating consistent text to another difficult problem (e.g. information extraction (IE) or NLI). 
Given a strong IE system or a structured representation of the source information, it is possible to dramatically improve the consistency of generated text \citep{zhang-etal-2020-optimizing, Tian2019StickingTT}, but such resources are only available in a narrow subset of domains.

\begin{table}[t]
    \centering
    \footnotesize
    \begin{tabular}{ll}
      \toprule
       Method&Text\\
      \midrule
      Source &...The PSNI said the tablets were ``as yet\\
            &unidentified'' but warned of the ``potential\\
            &dangers'' they posed...\\
      BART  &A 17-year-old boy has been charged after\\
            &a teenager was taken ill after taking what\\
            &police have described as \emph{\textcolor{red}{``potentially}}\\
            &\emph{\textcolor{red}{lethal'' ecstasy tablets}}.\\
      \sys  &A 17-year-old teenager has been charged\\
            &with drugs offences after a teenager was\\
            &treated in hospital after taking what police\\
            &described as an ``unidentified'' drug.\\
    \bottomrule
    \end{tabular}
    \caption{An example of hallucination. Inconsistent words are highlighted in \emph{\textcolor{red}{red italic}} fonts. In this case, \sys\ corrects the inconsistent detail in the BART output.}
    \label{tab:fig1}
    \vspace{-3mm}
\end{table}

We propose a different approach for generating more consistent summaries.  It is based on the observation that today's abstractive summarizers are often trained on target summaries that contain statements unsupported by the source text \cite{matsumaru-etal-2020-improving}.  This disconnect arises because the training datasets are acquired from noisy ``silver'' sources in order to scale, e.g. treating a news headline as a summary of its article or an encyclopedia entry as a summary of a portion of its references.
We conjecture that a model optimized for likelihood and trained on target summaries containing unsupported statements will have a strong tendency to hallucinate information rather than say something less ``likely,'' but supported (\secref{sec:motiv}). 
Further, common automatic evaluation metrics like ROUGE \emph{reward} lexical similarity significantly more than consistency, preferring hallucinated lexically similar summaries to completely consistent lexically different ones. 

Our method, called \sys, is a novel decoding algorithm that constrains beam search to only consider predicted tokens
that are likely to be supported by the source text. \sys estimates which tokens are likely supported using simple but effective heuristics based on the model's confidence and attention distribution, and word frequency.  When \sys reaches a state where no supported token can be generated, it backtracks the search. It can also opt-out from generating a summary at all, rather than produce one expected to be hallucinated.
We show how \sys\ significantly improves consistency on two abstractive summarization datasets with only a small decrease in fluency, measured using careful human evaluations.

To test \sys on diverse domains, we also develop a new abstractive summarization dataset called Scientific Concept Description (SCD). Inspired by the WikiSum~\citep{Liu2018GeneratingWB} dataset, SCD uses Wikipedia descriptions as the target summaries and the referenced papers as the source documents, detailed in (\secref{sec:datasets}). SCD is motivated by the goal of automatically generating a high-quality encyclopedia for the long tail of scientific concepts described in papers, and presents a challenging workload for abstractive summarization.  It comes with a total of 60k samples of scientific concepts and 118k corresponding paper identifiers, with full text for 8k of the papers.

We make the following contributions:
\begin{compactenum}
\item We analyze the relationship between hallucination and training on targets that are not fully supported by the source. %
\item We introduce \sys, a decoding algorithm that improves generation consistency by constraining beam search to focus on input-supported tokens. It improves consistency by an average of 68\% in two abstractive summarization datasets at the expense of a minor decrease to fluency.
\item We introduce Scientific Concept Description, a challenging new abstractive summarization task, and release a dataset.

\end{compactenum}

The SCD dataset, along with our code, trained models, and human evaluations, is available at \url{https://github.com/allenai/pinocchio}.

\section{Related work}
\label{sec:related}

Pretrained language models have recently taken the top spots on summarization leaderboards \citep{fabbri2020summeval, huang-etal-2020-achieved}. 
This includes models like BART \citep{lewis-etal-2020-bart}, PEGASUS \citep{Zhang2020PEGASUSPW}, and UniLM \citep{Dong2019UnifiedLM}. 
In a recent large scale evaluation of summarization models, \citet{fabbri2020summeval} found BART and PEGASUS to be the top performing models. 
We choose to focus on BART in this work. %

It is widely known that SOTA summarization models tend to hallucinate facts \citep{maynez-etal-2020-faithfulness}, and the most closely related works to ours are those on factual summarization.
However, we avoid the term ``factuality'' and instead use ``consistency'' to denote that the generated summary is supported by the input text.
As noted in~\citet{maynez-etal-2020-faithfulness}, a summary could be hallucinated but still be factually correct.
In this work, we aim to improve consistency and reduce hallucinations, which indirectly improves factuality, without directly optimizing for it.%

Prior works attempt to improve consistency by correcting already-generated summaries \citep{dong-etal-2020-multi-fact, Zhu2020EnhancingFC}, using a knowledge graph \citep{Zhu2020EnhancingFC}, filtering training data \citep{nan-etal-2021-entity}, constraining generation with keywords \citep{Mao2020ConstrainedAS}, using NLI models \citep{Barrantes2020AdversarialNF, Mishra2020LookingBS}, among others. Some have focused on the data-to-text setting, which presupposes structured input \citep{Tian2019StickingTT, wang-etal-2020-towards}. 
Some works control the extractiveness of generations \citep{DBLP:conf/aaai/SongWFL020}. 
There have also been multiple works on automatically measuring consistency~\citep{durmus-etal-2020-feqa, kryscinski-etal-2020-evaluating, wang-etal-2020-asking}. 
\citet{matsubara-singh-2020-citations} noted that hallucinations come from a source-target discrepancy, where many training targets are not fully supported by their source text, and suggested to address it by removing samples with unsupported summaries.  We extend their empirical findings with similar measurements on three additional datasets, conjecture that hallucination is unavoidable in such settings, and provide evidence for the conjecture in terms of the lexical statistics of output summaries. %

We use beam search for decoding, which has become standard practice for neural seq2seq models~\citep{Graves2012SequenceTW, Sutskever2014SequenceTS}.
Our approach can be viewed as a version of constrained decoding~\citep{hokamp-liu-2017-lexically} but with dynamically identified constraints and the ability to backtrack.
Our constraints come from various model internal signals that indicate attribution to the source text. 
One such signal is entropy, where \citet{xu-etal-2020-understanding-neural} found that low next token entropy indicates the model is copying.
Unlike previous work, we do not attempt to imbue models with a new level of textual understanding, but rather show that we can improve consistency of generated text using simple signals based on model internals.

\begin{figure}[t]
    \centering
    \includegraphics[width=0.9\columnwidth]{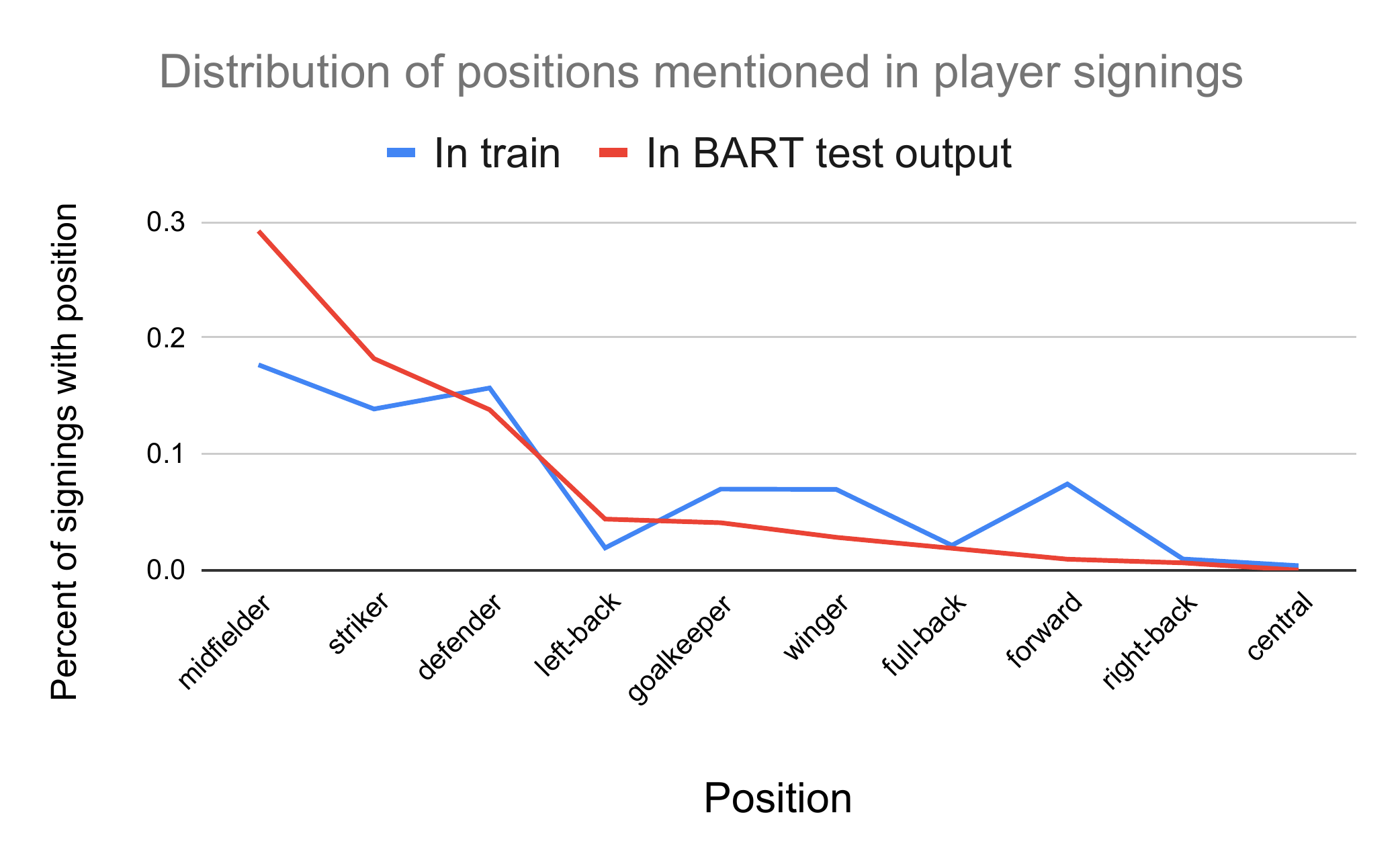}
    \caption{Distribution of positions in summaries about player signings in train vs. BART output.  BART output is more peaked at positions more common in train, suggesting BART defaults to these when no position is supported by the source.}
    \label{fig:peaked}
    \vspace{-5mm}
\end{figure}

\section{Why Do Models Generate Inconsistent Summaries?}
\label{sec:motiv}

In this section, we analyze why models generate inconsistent summaries.
Here, we use the definition of {\em consistent} from~\citet{fabbri2020summeval}, i.e., the factual alignment between the summary and the summarized source.

We hypothesize that there are two factors that contribute to inconsistency: 1) the maximum likelihood training and generation strategy used in summarization models, and 2) imperfect training datasets that contain many instances where the target is difficult or impossible to deduce from the source. 
Specifically, we conjecture that in the presence of these two factors, models are guaranteed to hallucinate because they either 1) default to a background distribution of the most common relevant terms during generation or 2) learn spurious correlations between the source and target texts. 
In either case, the model generates text that is often inconsistent with the inputs.

We present our analysis in terms of a motivating example below, and provide empirical support for it in  Sec.~\ref{sec:qual}. 
The analysis inspires the design of the \sys\ method in Sec.~\ref{sec:method}.

\subsection{Motivating example}
\label{sec:motivating-example}

Consider the target summary of an article about a team signing a football player, from XSUM:

{\tt \footnotesize 'League Two club Cheltenham Town have signed Hibernian striker Brian Graham on a free transfer.'}

Many of the details in this summary are difficult for a model to predict because they are {\em not} supported directly by the input passage.\footnote{The full input passage summarized in this example is in App. C.2.}  
For example, the player's first name (``Brian'') and position (``striker''), and the lack of signing fee (``free transfer'') are nowhere mentioned. 
This mismatch between typical summary fields and the text available in the input passage is not restricted to summaries about player signings, but is more generally observed across a variety of article types in XSUM and also our new SCD data set.

Achieving a high likelihood on the training dataset requires that the trained models output the aforementioned fields anyway: e.g., in summaries of player signings, from a sample of 43 summaries, 100\% mention the player's full name, 88\% the player's position, 78\% the length of the signing, etc, even though they are often not supported in the source. %
As a result, the BART summarizer outputs the following summary for the example:

{\tt \footnotesize 'League Two side Cheltenham Town have signed Hibernian midfielder Scott Graham on loan until the end of the season.'}

This summary begins nearly identically to the target, but then outputs the three field values incorrectly (first name, position, and length of the contract). 

The errors make sense when you consider the model's calculus for choosing a summary. Consider a single field that can be present or absent in a summary, and make the simplifying assumption\footnote{Note that the \sys~{\em method} (Sec.~\ref{sec:method}) does not depend on this assumption, it is only used here for intuition and ease of analysis.} that the probability of the most-likely summary with a field value is strictly monotonic in the probability of the field value (see App. B for formal details). In that case, a model that maximizes likelihood will output the field if and only if its best guess of the field value is more probable than the field's absence.  In practice, the probability of field absence is often low because training summaries of certain topics reliably cover certain fields, and the best guess probabilities are often higher because the model can do {\em some} inference to narrow the choice set to a limited and typically peaked distribution (e.g., to a small number of football player positions).  Thus, hallucinating a best guess is often preferred by the model---even, in some cases, when the model estimates that the guess is less likely than chance to be correct. In the example, since the estimated probability that the player is a ``midfielder'' is relatively high (``midfielder'' is relatively common, shown in Fig. \ref{fig:peaked}), and position going unmentioned is rare (about 12\% of the time), the model chooses to incorrectly output ``midfielder.''

Of course, the assumptions in our analysis may not always hold, and hallucination is likely more complex than the single phenomenon analyzed here.  But our approach, motivated by the above conjecture, can improve the consistency of summaries in practice.  Further, in Section \ref{sec:qual} we validate two aspects of our analysis empirically, showing that ground truth training summaries for abstractive summarization do contain unsupported statements, and that summarizers do disproportionately produce more common terms in their output.

\section{\sys: Constraining Beam Search to Improve Consistency}
\label{sec:method}

Inspired by the previous analysis, we introduce \sys; a modification to standard beam search for \textbf{supported-decoding} (Alg.~\ref{alg:supported}). 

\begin{algorithm}[t]
    \footnotesize
    \SetAlgoLined
    \KwIn{beam size $B$, generative model $M$, consistency function $f_c$, vocab $V$, maximumly allowed backtrack count $N$}
    priority queue $PQ$ = ["<start>"]*B\;
    completed generations $CG$ = \{\}\;
    rejected paths $R$ = \{\}\;
    backtrack count $\eta$ = 0;
    
\While{|CG| < B} {
 $C := \{x + v : x \in PQ, v \in V\} - R$\;
 $T :=$ top 2$B$ items of $C$ scored by $M$\;
 $R := R \cup \{d \in T : f_c(M, d) = 0\}$\;
 \If{$T-R == \emptyset$}{
   \If{$\eta \ge N$}{
    \tcp{Stop Generation}
    {\bf return} \{\}\;
   }
  $R := R \cup \{{\rm x[:-1]} : x \in T$\}\;
  $PQ := \{{\rm x[:-1]} : x \in PQ\}$\;
  $\eta := \eta + 1$\;
  {\rm \bf continue};
 }
 $T := T - R$\;
 $PQ :=$ top $B$ elements of $T$ according to $M$ not ending in "<end>"\;
 $CG := CG \cup \{d \in T : d$ scores higher than min in $PQ$ and ends in "<end>"\}\;
 }

{\bf return} top-ranked element of $CG$\;
\caption{Supported-decoding}
\label{alg:supported}
\end{algorithm}

Beam search for text generation typically works by adding to a small set of candidate generations one token at a time, keeping the top $B$ generations according to model-predicted likelihood after each prediction timestep.
After \texttt{<end>} has been predicted in $B$ beams, those $B$ candidates are rescored with a length penalty \citep{Wu2016GooglesNM}, and the best one is chosen as the final output.  \sys 
differs from regular beam search only in its use of the set R, which holds a set of disallowed generation paths; if R is always empty, Alg.~\ref{alg:supported} simplifies to standard beam search. 
\sys\ modifies the model predicted token scores to avoid inconsistent predictions.

\begin{table*}[t]
  \centering
  \resizebox{0.8\linewidth}{!}{
  \begin{threeparttable}
      \renewcommand{\arraystretch}{1.25}
      \begin{tabular}{llcccccc}
        \toprule
         \textbf{Method}&\textbf{Dataset}&\textbf{\% Cons.=5}&\textbf{\% Cons.= 4/5}&\textbf{Cons.}&\textbf{Flue.}&\textbf{Rele.}&\textbf{Cohe.}\\
        \midrule
        BART (n=282)& XSUM & 0.287 & 0.709 & 3.908 & \textbf{4.794} & 4.887 & -\\
        \sys\ (n=211)& XSUM & 0.422 & 0.82 & \textbf{4.19} & 4.649 & 4.886 & -\\
        BART (n=268)&SCD & 0.209 & 0.552 & 3.612 & \textbf{4.537} & \textbf{4.925} & 4.619\\
        \sys\ (n=207)&SCD & 0.396 & 0.768 & \textbf{4.082} & 4.338 & 4.816 & 4.585\\
      \bottomrule
      \end{tabular}
  \end{threeparttable}
  }
  \caption{Human evaluation of models. \sys\ improves consistency significantly, while decreasing fluency slightly. For the 4 evaluation metrics, significant (Mann–Whitney U test, p<0.01) differences are bolded. Cons.=Consistency, Flue.=Fluency, Rele.=Relevance, Cohe.=Coherence.  For each row, n denotes the number of examples output, which is lower for \sys\ than for BART because \sys\ elects to skip certain cases.}
  \label{tab:human_eval}
  \vspace{-3mm}
\end{table*}

In particular, \sys\ applies a function $f_c(\text{model state}, \text{candidate next generation})$ to the predicted likelihood of the top predicted tokens. 
If all top predicted tokens for a given timestep are inconsistent according to $f_c$, \sys\ backtracks by removing the last predicted token from each beam, and predicts again without the ability to predict the removed tokens. 
The number of times this backtracking occurs $\eta$, combined with the average entropy of the token predictions in the final output is a good indicator of whether the model succeeded in producing a good summary or not. 
Thus, we eliminate generations with multiple backtracks (e.g., $\eta>2$) and high entropy, as well as individual sentences with high entropy (>2.75) from multi-sentence outputs.%

Within this framework, we present an instantiation of $f_c$ based on a set of carefully curated heuristics, determining if a token is allowed to be predicted or not.

The function $f_c$ consists of a series of binary checks, which take into account both model internals as well as language features.  If any of the checks succeeds, $f_c$ is 1 and the model continues generating, but if all of the checks fail $f_c=0$ and the model disallows the generation path. 
First, we consider the {\em model confidence} for the current prediction---based on the intuition that a low entropy of the token prediction probability distribution corresponds to more certain, and potentially more correct, predictions.
Second, we keep track of the source text with high {\em attention} scores during the generation process: when the attended texts are semantically or lexically similar to the token to be generated, that suggests that the token may be supported by the source.  Third, \sys\ also allows tokens that are especially {\em common} (such as stopwords), as we expect these are less likely to be hallucinations. 
We develop a total of 8 different binary functions within the three categories above (details in \secref{subsec:fc-details}). 

The heuristics do not require additional training steps, and all the associated thresholds or hyperparameters were determined by manual inspection on a small number of samples (e.g., n=20) from each dataset. Different from prior work~\citet{matsubara-singh-2020-citations}, this non-machine learning approach is based on scrutiny of the model generation process. It is easy to execute and more explainable compared to black-box models.

\section{Tasks and Datasets}
\label{sec:datasets}

We evaluate \sys on two distinct summarization tasks: news summarization (XSUM and CNN~/~Daily Mail) and scientific concept description (the newly proposed SCD dataset). 

\subsection{News Summarization}

\paragraph{XSUM} \citep{narayan-etal-2018-dont} is a popular abstractive news summarization dataset.
XSUM is a challenging dataset; the source text frequently does not entail the target text, the target task is not exactly summarization (XSUM is closer to headline generation than summarization), and data is noisy (e.g. there are articles in another language, Welsh). 
Challenges aside, XSUM is highly regular, as mentioned in Sec.~\ref{sec:motiv}.
Although this seems to make the task easier, a strong pattern matcher will reproduce dataset patterns (see Appendix~\ref{sec:patterns} and Tab.~\ref{tab:patterns} for example patterns), \emph{whether or not} it is able to fill in all the details in the pattern correctly.

\paragraph{CNN~/~Daily Mail Dataset} ~\cite{nallapati2016abstractive} is another commonly used dataset for news summarization. 
Different from XSUM, the summaries are relatively longer (one sentence vs more than 2 sentences) and are considered to be nearly extractive (see \citet{sharma-etal-2019-bigpatent} and our results in Tab.~\ref{table:abstractiveness}) as the summaries are based on summary bullets from the original news article.

\subsection{Scientific Concept Description}
\label{subsec:scd}
We introduce the novel task of {\em scientific concept description} (SCD): automatically generating a brief description of a scientific concept, given the concept name and some papers discussing the concept. Test data has been manually evaluated to ensure quality. 

\paragraph{SCD training corpus}
Training an SCD system requires a large set of ground-truth descriptions.  
Inspired by the WikiSum dataset \citep{Liu2018GeneratingWB}, we construct our training set using Wikipedia intro sections\footnote{Specifically, we use the first section for the concept, and also include sections with definitional headers (Introduction, Definition, Uses, Description, Function, Overview).} as the target descriptions,\footnote{English Wikipedia 4/1/20 dump processed with \url{https://github.com/spencermountain/dumpster-dive}} with the papers cited in each description as source text. 
To remove intractable examples, we filter out those with lower than 0.15 ROUGE-1 recall between the cited papers and the target Wikipedia description.
The dataset is split into train/dev/test with 47570/5989/5839 examples.
Examples have 2.4 source documents with a total of 319 sentences on average and target descriptions averaging 6 sentences each. 
We are able to extract body text for \textasciitilde57\% of the cited papers, and use just the titles and abstracts of the remainder.%

\paragraph{Manually-evaluated SCD test corpus}
The motivating use case for the SCD task is automatically generating a high-quality encyclopedia for the long tail of scientific knowledge presented in papers.  
As a result, we construct a second test set of SCD evaluation examples not from Wikipedia, but instead from a much broader set of scientific concepts mined from computer science papers using ForeCite \citep{King2020HighPrecisionEO}.  
This set lacks target descriptions, so it requires manual evaluation.

Training on surrogate data that differs somewhat from the intended use case but can be obtained at scale is common in summarization research (e.g. abstracts as paper summaries \citep{Cohan2018ADA}; headlines as news summaries \citep{narayan-etal-2018-dont}). 
In our case there are two major discrepancies between train and test: the textual domain (train is mostly biomedical, test is largely computer science), and the level of supporting text (the Wikipedia-cited training inputs often have less support for the concept description than the ForeCite-mined test inputs do, as ForeCite pairs concepts with their likely introducing paper(s)).

\section{Experiments}
\label{sec:quant}

\begin{table}[t]
  \centering
  \resizebox{1.\linewidth}{!}{
  \begin{threeparttable}
      \renewcommand{\arraystretch}{1.25}
      \begin{tabular}{llcccc}
          \toprule
          \textbf{Method} & \textbf{Dataset} & \textbf{\# Samples} & \textbf{R1} & \textbf{R2} &\textbf{RL}\\
          \midrule
          BART    &XSUM & 11333 & 0.444 & 0.210 & 0.354     \\
          BART*   &XSUM & 8345\tnote{1} & 0.442 & 0.207 & 0.349 \\
          \sys\   &XSUM & 8345  & 0.431 & 0.196 & 0.338     \\
          \cdashline{1-6}[.4pt/1pt]
          BART    &SCD  & 5839  & 0.380 & 0.167 & 0.270    \\
          BART*   &SCD  & 2335  & 0.398 & 0.189 & 0.291    \\
          \sys\   &SCD  & 2335  & 0.391 & 0.181 & 0.284     \\
          \cdashline{1-6}[.4pt/1pt]                    
          BART    &CNN/DM  & 10990  & 0.438 & 0.209 & 0.372   \\
           BART*    &CNN/DM  & 10943  & 0.438 & 0.209 & 0.372   \\
          \sys\   &CNN/DM  & 10943  & 0.438 & 0.209 & 0.372          \\
          \bottomrule
          \end{tabular}
          \begin{tablenotes}
            \item[1] Because \sys\ can elect to skip in certain cases, we report two scores for BART model outputs: for all test samples, and for the samples where \sys\ generates results. 
          \end{tablenotes}
  \end{threeparttable}
  }
  \caption{Rouge scores on different datasets with and without using \sys. Datasets with higher abstractiveness (e.g., XSUM and SCD) may suffer from higher ROUGE drops when \sys is used.}
  \label{tab:auto_eval}
  \vspace{-4mm}
\end{table}

\subsection{Metrics}
\label{subsec:metrics}

We rely on human evaluation, as current automatic metrics are unreliable for evaluating factuality (see \secref{subsec:auto_metrics}). 
We are not targeting ROUGE metrics \citep{Lin2004ROUGEAP}, but present them for completeness.\footnote{\url{https://github.com/Yale-LILY/SummEval}}

For human evaluation, we use standard dimensions of consistency (does the source entail the target?), fluency (is the target grammatical, understandable English?), relevance (does the target contain important information for understanding the source?), and coherence (do the sentences flow together coherently?)\footnote{Coherence not used on XSUM as targets are 1 sentence}, with definitions adapted slightly from \citep{fabbri2020summeval} via calibration with our annotators. 
We also decided to rate consistency and fluency on a five-point 1-5 scale, but relevance and coherence on a coarser three-point 1,3,5 scale. 
See App. \ref{sec:app_anno} for annotation guidelines.

\subsection{Manual evaluation}
\label{subsec:manual}

In Tab.~\ref{tab:human_eval}, we report manual evaluation results, with each example annotated by one annotator (inter-annotator agreement is reported in Section \ref{subsec:iaa}). 
\sys\ improves overall consistency. 
Expressing the results in terms of precision and recall, treating perfectly consistent output (i.e., a consistency score of 5) as a true positive, Table \ref{tab:human_eval} shows that \sys\ improves precision by 68\% on average (47\% on XSUM, and 89\% on SCD) without hurting recall, yielding an F1 improvement from 0.209 to 0.345 and 0.287 to 0.361 on SCD and XSUM respectively.  The improvements in consistency arise from two cases: first, when \sys\ produces output, it is rated more consistent than BART on 44\% and 24\% of the examples from SCD and XSUM respectively, whereas BART is more consistent for only 16\% and 13\% (in the remaining cases, the two systems are equallly consistent). 
Second, on the examples where \sys\ produces no output, BART's output is tends to be less factually consistent than its average, scoring 0.30 and 0.44 points lower (on the 5-point consistency scale) than its average for SCD and XSUM respectively.  

\begin{table}[t]
    \centering
    \footnotesize
    \begin{tabular}{lccccccc}
      \toprule
       \textbf{Metric}&\textbf{Dataset}&\textbf{Cons.}&\textbf{Flue.}&\textbf{Rele.}&\textbf{Cohe.}\\
      \midrule
      tau &XSUM & 0.60 & 0.84 & - & -\\
      exact &XSUM & 0.66 & 0.89 & 0.96 & -\\
      compare &XSUM & 0.69 & 0.82 & 1.0 & -\\
      compare\textasciitilde &XSUM & 1.0 & 1.0 & 1.0 & -\\
      tau &SCD & 0.55 & 0.43 & 0.20 & 0.53\\
      exact &SCD & 0.55 & 0.69 & 0.93 & 0.80\\
      compare &SCD & 0.67 & 0.64 & 0.86 & 0.64\\
      compare\textasciitilde &SCD & 0.95 & 0.98 & 1.0 & 0.98\\
    \bottomrule
    \end{tabular}
    \caption{Mean agreement metrics between all pairs of annotators. tau=Kendall's tau, exact=exact agreement, see \secref{subsec:iaa} for compare and compare\textasciitilde. The very low/null correlation values are due to low variance in relevance.}
    \label{tab:iaa}
    \vspace{-3mm}
\end{table}

We see that \sys\ \emph{does} reduce fluency with respect to the base BART model, and further that the sentence level entropy filter applied in \sys\ sometimes removes the key first sentence that defines the entity in SCD, resulting in a decrease in relevance.  Pretrained language models are capable of producing incredibly fluent text and prior work on steering them over-optimizes for maximizing the highest likelihood output~\cite{Subramani2019CanUL, Subramani2020DiscoveringUS}.
As a result, steering them away from their highest likelihood output as \sys\ does is bound to reduce fluency.
Our results suggest that some of this fluency is coming at the cost of factual consistency, as the model has learned how to follow patterns to produce plausible sentences, but not necessarily while sticking to the source text (see \secref{sec:motiv} and Appendix \secref{sec:patterns}).

\subsection{Automatic evaluation}
\label{subsec:rouge}

For completeness, we report ROUGE 1, 2 and L (Tab.~\ref{tab:auto_eval}), for the two tasks along with results on the CNN/Daily Mail dataset~\cite{Hermann2015TeachingMT} for reference.
We note that \sys\ elects not to generate for a much higher portion of samples in SCD. This can be partially explained by the abstractiveness of these datasets, which we detail in Section ~\ref{sec:empirical-analysis}.  
For the examples where \sys\ generates, \sys\ lowers ROUGE moderately for the two abstractive datasets compared to BART, and by somewhat more on XSUM than on SCD. We analyze the ROUGE drop in XSUM in Appendix \secref{sec:patterns}). By contrast, \sys\ fires only rarely for the extractive CNN/DM dataset, and therefore its ROUGE scores are unchanged from BART.

\begin{table}[t]
    \centering
    \footnotesize
    \begin{tabular}{lcc}
        \toprule
        \textbf{Metric}&\textbf{FactCC}&\textbf{FEQA}\\
        \midrule
        tau & -0.02 & 0.233\\
        compare$\neq$ & 0.528 & 0.585\\
        mean/$\sigma$ pairwise ties & 1.354/1.464 & 0.108/0.096\\
        mean/$\sigma$ pairwise not ties & 1.699/1.518 & 0.113/0.1\\
        \bottomrule
    \end{tabular}
    \caption{Agreement between automated metrics and our annotations.  tau represents Kendall's tau, compare$\neq$ denotes agreement with the annotator on which model is better, \emph{when the annotator did not rate the models as equivalent}, "Mean/$\sigma$ pairwise ties" gives the mean/std of absolute value of difference between the metric's rating for each model, for pairs where the annotator rated the models as the same, and "Mean/$\sigma$ pairwise not ties" is the same but for pairs where the annotator rated the models as different. A well-calibrated metric should have mean near zero and low standard deviation when the models are annotated as equivalent.  We find the automated metrics exhibit low agreement with our annotators.}
    \label{tab:auto_agreement}
    \vspace{-3mm}
\end{table}

\begin{table*}[t]
    \centering
    \resizebox{1.\linewidth}{!}{
    \begin{threeparttable}
        \renewcommand{\arraystretch}{1.25}
        \begin{tabular}{cccccrccccc}
            \toprule
            \multicolumn{1}{l}{}    & \multicolumn{5}{c}{\textbf{Dataset Abstractiveness}}& \phantom{\small{1}} & \multicolumn{2}{c}{\textbf{Human Annotated Unsupported Words}} &\phantom{\small{1}}  & \textbf{BART+\sys} \\
            \cmidrule{2-6} \cmidrule{8-9} \cmidrule{11-11}
            \textbf{Dataset}        & 1-gram   & 2-gram   & 3-gram   & 4-gram   & Avg.     &  & \% Unsupported Words  & IAA - Cohen $\kappa$                    &  & Avg. $\eta$ per Successful Generation                     \\
            \midrule
            \textbf{CNN/DM}\tnote{1} & 17.18\%  & 58.44\%  & 78.06\%  & 86.71\%  & 60.10\% &  & 1.57\%                 & 0.571                          &  & 0.0003   \\
            \textbf{XSUM}           & 49.88\%  & 89.65\%  & 98.13\%  & 99.60\%  & 84.31\% &  & 17.78\%                & 0.728                          &  & 0.1541   \\
            \textbf{SCD}            & 60.16\%  & 88.97\%  & 96.81\%  & 98.61\%  & 86.14\% &  & 23.84\%             & 0.414                          &  & 0.2300   \\
            \bottomrule
            \end{tabular}
            \begin{tablenotes}
                \item[1] We report the scores for the CNN / Daily Mail dataset~\cite{see2017get,HermannKGEKSB15} for comparison because it is highly extractive.
            \end{tablenotes}
    \end{threeparttable}
    }
\caption{Analysis of the abstractiveness of three summarization datasets.  The abstractive XSUM and SCD datasets contain a substantial fraction of unsupported words,  measured in terms of either automated n-gram overlap measures or manual annotation. BART+\sys\ performs more backtracks $\eta$ on more abstractive datasets. 
}
\label{table:abstractiveness}
\vspace{-4mm}
\end{table*}

\begin{figure*}[t]
    \centering
    \includegraphics[width=0.8\linewidth]{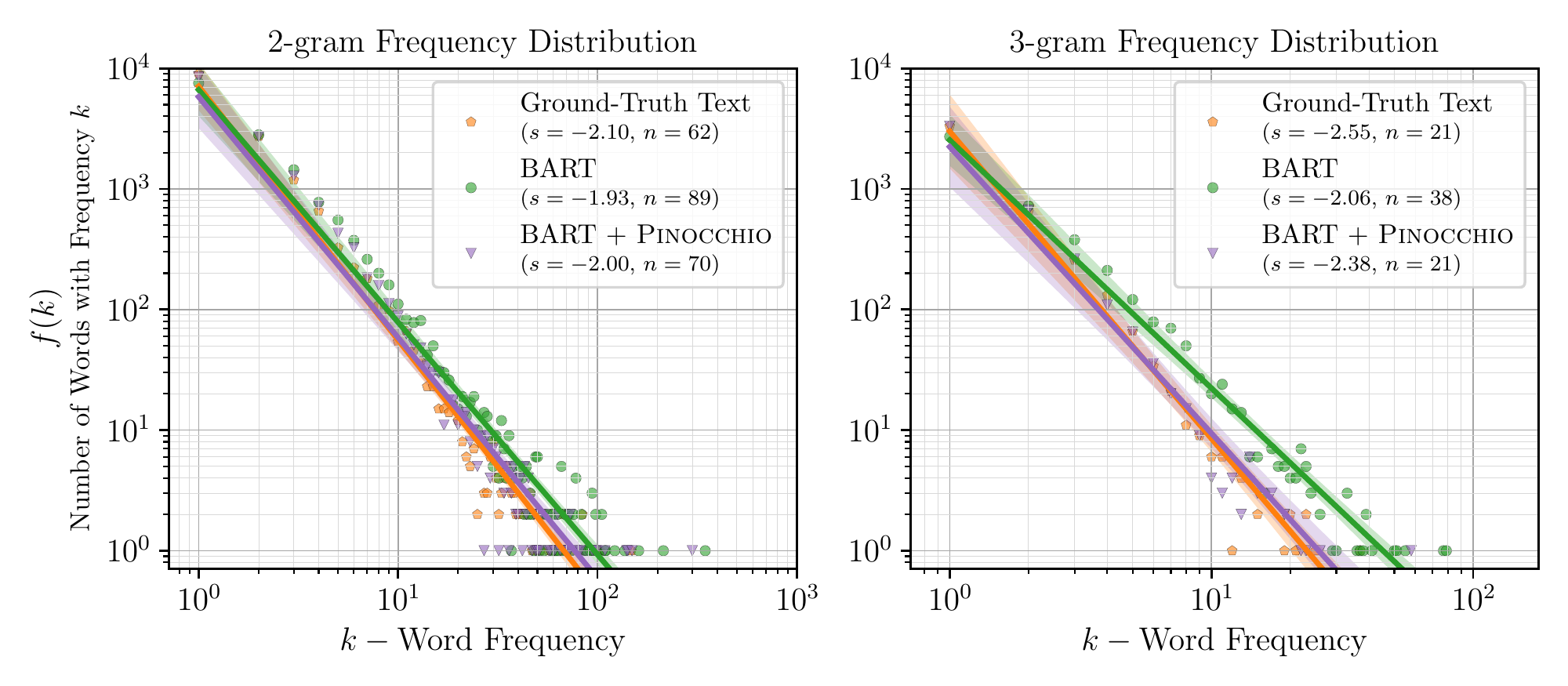}
    \caption{Comparing the n-gram frequency distribution on the XSUM Dataset for generated, versus ground truth sources. The default BART model outputs (in green) over-represent frequent n-grams (bottom right of the distribution), but \sys\ is closer to the ground-truth. Results in the SCD dataset are similar. The slope of the linear fits for ground-truth text and BART generations are significantly different ($p\ll 0.05$, ANCOVA) while those between ground-truth and BART + \sys generations are not ($p>0.05$) for both 2-gram and 3-grams.}
    \label{fig:power-law}
    \vspace{-3mm}
\end{figure*}

\subsection{Comparison against existing correctors and factuality metrics}
\label{subsec:auto_metrics}

We also compare with three recent methods for automatically correcting summaries or measuring their factuality.  
Here we evaluate on XSUM, which we expect to be more suitable for these methods (each were evaluated on XSUM in previous work, whereas SCD is out of domain).
First, we compare against \citet{Zhu2020EnhancingFC}, a recent seq2seq fact corrector (FC) that incorporates OpenIE \citep{Angeli2015LeveragingLS} and knowledge graph embedding.  
We take the output of their strongest model (UniLM \citep{Dong2019UnifiedLM}+FC) on the XSUM test set and find that it changes only \textasciitilde5\% of examples, and that the net improvement rate of the changes is 15\% (see App. F for details).  
This corresponds to an improvement on <1\% of the full XSUM test set.  
By contrast, our experiments in the previous section show that \sys\ yields an improvement on \textasciitilde8.5\% of XSUM, more than a factor of eight higher.

Finally, we assess two representative automatic factuality metrics, FactCC \citep{kryscinski-etal-2020-evaluating} and FEQA \citep{durmus-etal-2020-feqa}. 
FactCC trains a <source, summary sentence> classifier; FEQA generates/answers questions from the summary, checking if answers are the same when using the source.
We find neither metric suitable for our highly abstractive setting; each has low agreement with our XSUM annotations (Tab.~\ref{tab:auto_agreement}), a result in line with a very recent evaluation of factuality measures \citep{Pagnoni2021UnderstandingFI}.

\vspace{-1mm}
\subsection{Inter-annotator agreement}
\label{subsec:iaa}

In Tab.~\ref{tab:iaa}, we report various inter-annotator agreement measures. 
We had three expert annotators, and the agreement stats are averaged between all pairs of annotators, on a set of 30 examples (15 from each model) from each dataset. 
For model comparison, the most important metrics are the ``compare'' metrics, which measure how often the annotators agree on which model's output is better for a given example. 
The ``compare'' metric is the fraction of examples for which the pair of annotators agree on which model's output is better or both say the outputs are equivalent. 
The ``compare\textasciitilde'' metric is similar but more lenient, as it only counts as disagreement the examples where one annotator says one model is better, and the other annotator says the opposite.
These kinds of strong disagreements are very rare in our data, suggesting that the relative comparisons between models in our experiments are reliable.

\section{Discussion}
\label{sec:qual}

\subsection{Empirical validation of the intuition motivating \sys}
\label{sec:empirical-analysis}

We now present two empirical analyses to verify the intuition sketched in Section \ref{sec:motiv}.  First, we verify our claim that the ground truth summaries in our data sets contain unsupported terms (Table~\ref{table:abstractiveness}).  We define \emph{Dataset Abstractiveness} as the ratio of n-grams that appear in the summary but not in the source text.  The two abstractive datasets (XSUM and SCD) show high abstractiveness, with approximately half or more of the terms in the summaries not appearing in the source.  Of course, a lack of lexical overlap could arise from summaries stating supported information but in different terms from the source.  Thus, we also manually examine twenty examples for XSUM and CNNDM and ten for SCD and measure the fraction that are not directly supported by the source.\footnote{This annotation task can be challenging and subjective especially for the SCD dataset, see appendix \secref{sec:annotation_difficulty} for details.}
This fraction is substantial (18-24\%) for the abstractive datasets, but much smaller (2\%) for the more extractive CNN/DM dataset.  Finally, $\eta$, the number of times our proposed method BART + \sys\ %
{\em backtracks}, which is a measure of how often the method estimates that generated tokens are unsupported, also correlates with the abstractiveness measures.

We also verify one expected consequence of our hypothesized mechanism of hallucination.
If indeed BART is defaulting to a background distribution of field values (based on frequency in the training summaries), then we would expect the more frequent training values to become even more probable in BART's output, as the model defaults to these as best guesses. We observe this effect for positions in player signings, as shown in Fig. \ref{fig:peaked}.  It is notable that while this distribution is more peaked, it is not entirely concentrated on the most-likely field value, suggesting that the model has learned spurious correlations that lead it to output other more rare field values, even when unsupported.

More generally, we also observe a similar bias across all n-grams; compared to the original ground truth summaries, the BART output tends to be less heavy-tailed, including disproportionately more of the high-likelihood n-grams.  We show this by plotting the n-gram frequency distributions (which follow a power law) on a log-log scale in Fig.~\ref{fig:power-law}.  The BART output generally has a less negative slope than the ground truth distribution on these plots. BART + \sys\ method results in a distribution that is closer to the ground truth for 2- and 3-grams.

\subsection{Errors analysis}
\label{subsec:error}
To provide insight into dominant error types, we sample 20 \sys\ generations from the SCD evaluation with inconsistent outputs, and identify three common error causes that each occur in \textasciitilde20\% of the samples: 1) Incorrect paraphrasing or omission of meaning-changing information (e.g. X has a long history of being used for Y vs. X is the model of choice for Y) 2) Incorrect treatment of entities as coreferent/synonymous 3) Difficulty with heavy mathematical notation. 

We also provide additional qualitative analysis on the generated outputs in Appendix \ref{sec:patterns}. We conclude that BART tends to exploit specific patterns in the dataset that contribute to its better ROUGE scores, but it fails to reliably apply commonsense or facts learned during training. Targeting these challenges in generative models is a promising future direction. 

\section{Conclusion}
\label{sec:conc}

In this work, we present \sys, a simple, no-additional-machine-learning required, method for %
reducing hallucination in generative encoder-decoder models. \sys\  provides a substantial lift in consistency, with only a small decrease in fluency. 
We analyze why existing summarizers hallucinate, showing that silver abstractive summarization datasets can contain unsupported target summaries, and presenting evidence for our conjecture that models that maximize likelihood trained on such data will tend to hallucinate. 
We also show that existing factuality metrics are insufficient, and further explore how patterns in the training dataset can produce misleading results on the test test. 
We also introduce the task of scientific concept description and release a Wikipedia-based dataset for it.

We would like to clearly acknowledge the limitations of our approach. 
\sys\ does not add new learned behavior to the model, using simple heuristics and single-step backtracking to steer the model towards more consistent output.  The heuristics have settings that require some adaptation for each data set, and while limited manual tuning was sufficient for the two data sets in our experiments, further experiments with additional data sets are necessary.  Further, preliminary experiments suggest that the settings that were effective for BART do not simply work out of the box for another summarizer, PEGASUS.  We also acknowledge that while \sys\ offers improvements in consistency, the results are still far from perfect, and thus the system is not suitable for certain applications.  We hope the approach and insights in this paper help spur further development of models that generate consistent text.

\paragraph{Acknowledgement}
The authors would like to thank Noah Smith and Kyle Lo for helpful comments.  This material is based upon work supported by  NSF Grant
OIA-2033558, NSF RAPID grant 2040196, ONR grant {N00014-18-1-2193}, and the Allen Institute for Artificial Intelligence (AI2).

\bibliography{anthology,custom}
\bibliographystyle{acl_natbib}

\cleardoublepage
\appendix
\label{sec:appendix}

\section{Example Full Text}
\label{sec:full_source}

\subsection{Example from Table 1}
\label{sec:figure1_example}

Police said the 14-year-old reported feeling unwell and required hospital treatment. He was later discharged from hospital and is recovering at home. The incident happened in Holywood, County Down, on Saturday. The PSNI said the tablets were "as yet unidentified" but warned of the "potential dangers" they posed. The 17-year-old, has been charged with possessing a Class A controlled drug with intent to supply; possessing a Class B controlled drug with intent to supply; possession of a Class A controlled drug; possession of a Class B controlled drug and supplying a Class A controlled drug. He is due to appear at Newtownards Youth Court on 14 February.

\subsection{Player signings}
\label{sec:signings_ex}

The 29-year-old Scot has signed a two-year contract with the Gloucestershire outfit.  Prior to joining Hibs in August 2016, Graham had spells at six other Scottish sides, including Dundee United, St Johnstone and Ross County.  He will be available for Saturday's league visit of Crawley Town, subject to receiving international clearance.  Find all the latest football transfers on our dedicated page.

\section{Mathematical Details of Hallucination Analysis}
\label{sec:app_motiv}

Formally, a summarization model is defined by a distribution $P(S | P)$ over output textual summaries $S$ conditioned on an input passage $P$.  We assume that the summarization system aims to maximize the probability of the summary $S$ given the text passage, i.e. it outputs $\argmax_S P(S | P)$.  While in practice (including in our experiments), summarization models use imperfect search procedures like beam search to find high-likelihood generations, and may rescore complete generations using factors other than likelihood (like length), in this analysis we ignore these details and assume the generator simply maximizes likelihood.  Analyzing the impact of more complex generation aspects is an item of future work.

Let $F(S)$ be a function denoting the value of a given ``field'' in the summary $S$, equal either to some string value or to $\emptyset$ if the field does not occur in $S$.  A ``field'' is a typical piece of information that is often mentioned in a summary of a given topic (e.g., participating teams, in a summary of a sporting event; or the university where an idea was developed, in a scientific concept description).  Then the model's distribution over a field value for a given passage is $P(F={\bf f} | P) = \sum_S P(F(S)={\bf f} | P)$.

Our analysis uses the following assumption:

Assumption {\bf A1}:
The model's most likely summary probability is strictly monotonic in the probability of its included field values.  That is, whenever:
\begin{equation}
    P(F = {\bf f} | P) > P(F = {\bf f'} |  P) 
\end{equation}
then
\begin{multline}
    \max_S P(S, F(S) = {\bf f} | P) > \\
      \max_S P(S, F(S) = {\bf f'} | P) 
\end{multline}

That is, when the model thinks a field value is more likely in a summary for a given passage, then it can find a more likely summary that uses that field value.  This assumption seems likely to hold often in practice (for example, we would expect that by simply swapping out a less likely field value in a summary for a more likely one, we would often arrive at a more probable summary).

The observation used in the analysis in Section \ref{sec:motiv} is then: given a passage $p$, a field $F$, and a summarization model $P(S | P)$, if assumption A1 holds, then a generator that maximizes likelihood will choose to output ${\bf f} = \argmax_{\bf f} P(F = {\bf f} | P)$ for the field's value (or omit the field, if ${\bf f}=\emptyset$).  This fact is straightforward from the definitions.

\section{Manual Examination of the Unsupported Dataset Samples}
\label{sec:annotation_difficulty}

Identifying parts of a summary that are not supported by the source document is a challenging annotation task. In this section, we explain how we formalize this task as a binary token tagging problem, and we show one example that illustrate the difficulty of annotation.

\subsection{Annotating unsupported words}

Naturally, words that appear only in the summary but not the source document tend to have a higher chance of being ``hallucinated'', and vice versa. Hence, we select such words from the summary, and the goal is try to identify whether the meaning of these words can be deduced from the source documents. Compared to the automated measurements, the manually inspected labels are considered to be a better approximation of the true abstractiveness of the dataset or the samples.

\subsection{One challenging example}

In practice, understanding the source document involves multiple (common sense) reasoning steps and subjective judgements.

Considering the following document text:

{\tt \footnotesize 'ABC of allergies: Venom allergy Stings from bees and wasps, the most common stinging insects in Britain, can cause severe allergic reactions, including anaphylaxis. Coroners’ data suggest that an average of four deaths from bee or wasp stings occur each year in the United Kingdom, but this is almost certainly an underestimate because venom anaphylaxis is not always recognised as the cause of death'}

For one sentence in the summary, we highlight the words that do not appear in the source in red:

{\tt \footnotesize 'The stings of most of these species (Bees) can be \textcolor{red}{quite painful}, and are \textcolor{red}{therefore keenly} avoided by many \textcolor{red}{people}.'}

The source text mentions several dangerous aspects of bee stings, but whether it can be concluded that they are avoided by many people (a plausible commonsense implication) is subjective to judge, and annotators often had differing opinions on these judgments.

\vspace{1mm} %
\section{\sys\ Details}

\subsection{Heuristics in $f_c$}
\label{subsec:fc-details}

We develop 8 binary checks that constitute the heuristics for $f_c$, which fall into three categories.  Two categories use model internals, \emph{model confidence} and \emph{source text attribution} for the predicted token. The third category uses language features, allowing generations that are {\em common words}.

\vspace{4mm} %
\vspace{4pt}
\noindent \textbf{Model confidence}
\begin{compactitem}
    \item entropy of next-token distribution < $\tau$ for a token in the top 2 predictions
    \item from the top 10 predicted next tokens, the number that match a top 5 attended-to piece of source text\footnote{All reference to ``top attended-to pieces of the source text'' means a max across locations in the source text across attention heads in the final layer of the decoder's cross-attention, and a 10-wordpiece window around the attended-to location.} is $>= \frac{1}{2}(10 - $the number that are stopwords$)$
\end{compactitem}
\textbf{Source text attribution}
\begin{compactitem}
    \item the most attended-to piece of the source text contains the predicted token
    \item 3 out of the top 5 attended-to pieces of the source text contain the predicted token %
    \item sum of the attention scores of the attended-to pieces of source text (out of the top 5) that contain the predicted token is greater than $\frac{1}{3}$ of the sum of the top 5 attention scores
    \item max cosine similarity between the embedding of the predicted token and that of any word in the top 5 attended-to pieces of source text is greater than 0.15 (and the word is not capitalized or a number word)*\footnote{Items marked with an asterisk * are optional.}
\end{compactitem}
\textbf{Common word}
\begin{compactitem}
    \item predicted token is a stopword*
    \item prediction matches\footnote{For all string matching, we lemmatize first.} one of the top 5 predictions of roberta-base\footnote{\url{https://huggingface.co/roberta-base}}%
\end{compactitem}
\vspace{4pt}

All of the components and hyperparameters above were determined via inspection on a small number of samples (e.g., n=20) from the XSUM and SCD dataset. In the subsequent sections we detail the configurations of the parameters on each dataset.

\subsection{XSUM modeling details}
\label{subsec:xsum_diff}

For configuration of \sys\ for XSUM, we set $\tau=1.0$ and do not use the optional stopword condition, in order to accommodate the highly abstractive nature of the XSUM dataset and attempt to prevent the use of stopwords in hallucinations. 

One other important detail is that XSUM has a surprising property with respect to first names. 
If a person appears in the source as ``Mr/Ms'' X, and also in the headline, they \emph{always} appear as <FIRST NAME> X in the headline. 
This leads to BART \emph{always} guessing the first name of a person, frequently incorrectly. 
Our $f_c$ often identifies the first name as unsupported, but because BART is essentially unable to predict anything other than a first name in this situation, it is unable to recover from this error. 
For this reason, when an unsupported token is identified as a name using spaCy \citep{spacy}, we deterministically replace it with Mr/Ms.\footnote{For real applications, we suggest using a gender neutral honorific, as gender is not possible to infer using first names}

\subsection{SCD modeling details}
\label{subsec:scd_model}

For SCD, the source consists of full papers and is too long to input to BART directly, so we train a separate BERT-based model to extractively rank chunks of the input text based on predicted ROUGE-L F1 score against the target text. 
This setup of ranking extractive chunks and then passing them to an abstractive model is similar to prior work on long text summarization \citep{liu-lapata-2019-hierarchical}. 
We pass the concept name/aliases and each chunk of text to rank to SciBERT-base \citep{beltagy-etal-2019-scibert}, with a final linear layer to predict the ROUGE-L score. We then finetune BART, with the ranked extractive chunks as source, again concatenated with the concept name/aliases. 
For inference, we also filter the chunks to those that include the concept name or an alias. 

\paragraph{Beam search parameters} 
We use standard parameters for the beam search of min\_length=5, max\_length=500, no\_repeat\_ngram\_size=3, length\_penalty=2.0, and num\_beams=6.

\paragraph{Extractive ranker for descriptions} 
The extractive ranker uses SciBERT\footnote{\url{https://huggingface.co/allenai/scibert_scivocab_uncased}}, followed by a linear layer, and is trained with MSE loss. We also use dropout of 0.1. We train on chunks containing three sentences, and use the average ROUGE-L as the label. To reduce the size of the training set, for each target description, we select the top 5 and bottom 5 chunks by ROUGE-L, and an additional 5 random chunks from the middle. We train for 3 epochs, with a batch size of 1, 8 gradient accumulation steps, and the AdamW \citep{Loshchilov2017FixingWD} optimizer, with weight decay 0.01, and a slanted triangular learning rate scheduler with peak learning rate 5e-5.

\subsection{Finetuning BART on descriptions} BART was finetuned with the standard settings,\footnote{\url{https://huggingface.co/facebook/bart-large/blob/main/config.json}} a batch size of 4 with 8 gradient accumulation steps, for 10 epochs, selecting the epoch 5 model based on validation loss. The same optimizer as above was used, with 500 warmup steps. The model was trained for 5.5 hours on 3 NVIDIA Quadro RTX 8000s. We additionally filter out examples that have a target length less than 150 characters, and examples where the source and target have less than 0.2 token overlap.

For configuration of \sys\ for SCD, we set $\tau$=0.75 and do not use the optional cosine similarity condition, to encourage more extractiveness.

\section{Annotation Instructions}
\label{sec:app_anno}

\begin{compactitem}
    \item Consistency
    \begin{compactitem}
        \item 1: completely made up
        \item 2: some phrases supported, but largely made up
        \item 3: some full details correct, but key details made up
        \item 4: minor details not fully supported (e.g. acronym wrong, location abstracted a bit wrong)
        \item 5: fully supported
        \item Other notes: An unresolved “it” should be assumed to refer to the main concept. If this makes it not factual, that counts against consistency, otherwise it counts against coherence.
    \end{compactitem}
    \item Coherence
    \begin{compactitem}
        \item 1: all sentences/phrases don’t make sense together
        \item 3: some sentences/phrases don’t make sentence together, separate from whether they are factual
        \item 5: no issues with how phrases/sentences are put together
    \end{compactitem}
    \item Fluency (at the sentence level)
    \begin{compactitem}
        \item 1: not fluent English to the point that it is impossible to understand/meaningless
        \item 2: not fluent English to the point that it is very hard to understand
        \item 3: semi fluent English (including major fluency errors resulting from copying source text), but still largely understandable
        \item 4: Mostly fluent English (including minor fluency errors resulting from source text), does not impact understanding
        \item 5: Fluent English
    \end{compactitem}
    \item Relevance
    \begin{compactitem}
        \item 1: off-topic
        \item 3: mostly on-topic or seems to be missing an actual statement of what the concept is (or for news, what the article is about)
        \item 5: on-topic and contains the key statement of what the concept is (or for news, what the article is about)
    \end{compactitem}
\end{compactitem}

\section{UniLM+FC Comparison Details}
\label{sec:app_fc}

Model output downloaded from \url{https://drive.google.com/file/d/1blmmJvniToN1yedoWUH3u0SNtXnMVDAs/view?usp=sharing} on 03/23/21. We consider an output ``changed'' by FC if it is not a prefix match for the original UniLM output, after lowercasing and removing spaces and apostrophes. Many FC-corrected examples seem to simply cutoff the end of the generated text. We choose to not count these as ``changed.'' There are 579 such cases. Given this criteria, FC changes 594 examples in the XSUM test set, and we sample 100 of these for evaluation. FC makes very minimal edits, so it is straightforward to identify whether the edit is an improvement or not. The net improvement is the number of increases in consistency minus the number of decreases in consistency.

\section{Patterns and Hallucination}
\label{sec:patterns}

\begin{table*}[ht]
    \footnotesize
    \begin{tabular}{llccc}
        \toprule
        \textbf{BART generation}&\textbf{Manual pattern}&\textbf{Train/val}&\textbf{Predicted}&\textbf{Consistent}\\
        \midrule
        A 70-year-old man who died after being&.*year-old.*who died.*named.*&47&4&0\\
        hit by a car in Monmouthshire has been\\
        named by police.\\
        \midrule
        Chinese businessman Dr Tony Xia has&.*Tony Xia.*Aston Villa.*|&7&1&0\\
        completed his £52m takeover of&.*Aston Villa.*Tony Xia.*\\
        Championship club Aston Villa.\\
        \midrule
        All pictures are copyrighted.&.*All pictures are copyrighted.*&44&4&4\\
        \midrule
        Forfar Athletic extended their lead at the&.*extended.*top.*points.*win.*&9&3&0\\
        top of Scottish League Two to five points&.*Forfar Athletic.*top of Scottish&10&1&0\\
        with a 3-0 win over Berwick Rangers.&League Two.*\\
        \bottomrule
  \end{tabular}
  \caption{Top-5 BART generations, by ROUGE-L gain over \sys\  (\#2 is excluded; it doesn't match an obvious pattern and is factually consistent). In all examples, BART clearly memorized training patterns and guesses the details in at least 3 (the 3rd output is memorized from noise in XSUM), which is not strongly penalized by ROUGE.}
  \label{tab:patterns}
  \vspace{-4mm}
\end{table*}

We provide additional discussions for some qualitative aspects of our results. 
First, we need to discuss the substantial drop in ROUGE on XSUM. 
As alluded to in \secref{sec:motiv}, we believe this is due to a pervasive regularity in the XSUM dataset, which BART is able to capture very well. 
In Tab.~\ref{tab:patterns}, we show the top examples sorted by ROUGE-L difference between BART and \sys, along with a hand-crafted regex matching the example, how many times it matches target outputs from the training and validation set, how many times it matches BART predictions on the test set, and how many of those predictions are completely factually consistent. 
Most of these examples straightforwardly map to patterns of text that occur in the training data. 
We also see that test set predictions matching these patterns are largely not consistent. 
As discussed in \secref{sec:motiv}, this is because BART assigns high likelihood to the general pattern, but guesses to fill in the details. 
Some of these patterns are straightforward to identify, but many are likely to be more complicated. 
Broadly speaking, XSUM contains a lot of regularity in the mapping between the source topic, phrases, and vocabulary used in the target summary.
BART exploits this, whereas
\sys\ steers the model away from the patterns, which are often not supported by the source text, which lowers ROUGE. 

A related question is if BART trained on XSUM applies facts learned during training correctly. 
Does it learn that Antonio Conte is the coach of the Italian football team, thus someone named ``Conte'' who coaches the Italian team is Antonio Conte? 
Or does it merely learn the first name most commonly associated with ``Conte'' in train is ``Antonio'', and so everyone named ``Conte'' is Antonio Conte? \footnote{Experiments with this example strongly suggest the latter.} 
It is difficult to assess this automatically, so we present an example of BART's tendency to guess world knowledge. 
We create one three-sentence source, ``Sometime last week, a fire burned down a <BUILDING>, killing a number of people. The fire took place in <LOCATION>. Investigators believe at least four people to be missing.'', filling in the blanks with three made up locations and three building types. 
BART produces plausible but inconsistent summaries. 
Nine out of nine outputs hallucinate the location, eight discuss arrests or hospitalizations, and three mention the police or fire service reporting the details of the situation. 
These characteristics are all due to biases present in the training data. 
Locations are often abstracted, reported fires often result in someone being arrested or hospitalized, and they are usually reported by authorities. 
We present this example as evidence that BART is not learning how to reliably apply commonsense and learned facts, but rather, is naively reproducing patterns and word associations.

\section{Comparing Generated Summaries with and without \sys }

In Tab.~\ref{table:generation-comp-1} and \ref{table:generation-comp-2}, we include example summaries generated with and without \sys. 
We additionally include the annotator ratings and their comments to illustrate how \sys\ improves the quality of the summaries.

\begin{table*}[ht]
    \resizebox{1.\linewidth}{!}{
    \begin{threeparttable}
        \renewcommand{\arraystretch}{1.15}
        \caption{Side-by-side comparison of the generated summaries with and without \sys \xspace -- Example 1 in XSUM.}
        \label{table:generation-comp-1}
        \begin{tabular}{p{0.15\linewidth}cccc}
        \toprule
        \textbf{Source} & 
        \multicolumn{4}{p{0.85\linewidth}}{Tourism NI said it expects a strategy to be in place by early next year. Janice Gault from the Hotels Federation told the BBC's Inside Business programme it was crucial for the industry. She said a "partnership" approach was essential. "I mean we've really urged people to get a strategy at sort of quite a high level so that everybody can buy into that," she said. "Hotels have probably spent about a billion pounds in the last decade and are set to spend more." Ms Gault said another big boom was expected in the hotel market which would probably generate another half a billion pounds. "The funny thing about the strategy is we still have the target, but we don't have the strategy. We only have one way to go and that's growth and the way for us to get that is to partnership," she added.} \\
        \midrule
        \textbf{Generation} &  
        \multicolumn{2}{c}{\textbf{BART}} &
        \multicolumn{2}{c}{\textbf{BART+\sys}} \\
        &
        \multicolumn{2}{p{0.4\linewidth}}{The Northern Ireland Hotels Federation has called on the Northern Ireland Executive to set out a strategy for growth in the hotel industry.} & \multicolumn{2}{p{0.4\linewidth}}{A new strategy for Northern Ireland's hotel industry has been urged by the Hotels Federation and Tourism NI as the industry is set for another big year.} \\
        \midrule
        \textbf{Ratings} &
        \multicolumn{2}{c}{Consistency: 3 \ Fluency: 5 \ Relevance: 5} &
        \multicolumn{2}{c}{Consistency: 5 \ Fluency: 5 \ Relevance: 5} \\
        \midrule
        \textbf{Annotator}\newline\textbf{Comment} &
        \multicolumn{2}{p{0.4\linewidth}}{\emph{Executive abstracted; didn't call on her; strategy is not to "grow" the industry}} \\
        \bottomrule
        \end{tabular}
    \end{threeparttable}
    }
\end{table*}

\begin{table*}[t]
    \resizebox{1.\linewidth}{!}{
    \begin{threeparttable}
        \renewcommand{\arraystretch}{1.15}
        \caption{Side-by-side comparison of the generated summaries with and without \sys \xspace -- Example 2 in XSUM.}
        \label{table:generation-comp-2}
        \begin{tabular}{p{0.15\linewidth}cccc}
        \toprule
        \textbf{Source} & 
        \multicolumn{4}{p{0.85\linewidth}}{The University and College Union says the 1.1\% rise offered by the universities is "an insult". But the Universities and Colleges Employers Association said the walkout was "disappointing given the very good pay offer". Unions representing university support staff are balloting on the offer, with strike action possible in the autumn. UCU says its members have suffered a real-terms pay cut of 14.\% since 2009 and complains the squeeze on staff salaries has come as university leaders enjoyed hefty increases. "A 1.1\% pay offer is an insult to hardworking staff, especially in light of the 5\% pay rise vice-chancellors have enjoyed while holding down staff pay," said general secretary Sally Hunt. "Industrial action which impacts on students is never taken lightly, but members feel that they have been left with no alternative. "If the employers wish to see a swift end to this dispute, and avoid further disruption, they need to come back to the table with a much-improved offer." Summer exams are still running at some universities, though many have finished. A spokesman for the employers anticipated only "minor impact and minimal student disruption". "Even for examinations which are still taking place at some higher education institutions, more than nine out of 10 report that a no to low impact is anticipated," said the spokesman. "We would like to see the UCU consulting its members on the final offer." The employers say the offer is "at, and, for some, beyond, a limit of affordability for higher education institutions and the very best offer that will be available this year". They maintain the weighting of the offer means the worst paid university staff will get a rise of more than 5\%. They say they have also offered talks on zero-hours contracts and on improving lower pay for female academics. But UCU says it rejected the 1.1\% offer as it was only a marginal improvement on the original 1\% on which it had balloted members. Ballots of university support staff represented by Unison and Unite are also under way on the improved offer, with both unions recommending it be rejected. Any action would take place during the autumn term, said a Unison spokeswoman. UCU is planning strike rallies in: Staff are also working to contract from Wednesday - refusing to set extra work, cover for absent colleagues or work overtime.
        } \\
        \midrule
        \textbf{Generation} &  
        \multicolumn{2}{c}{\textbf{BART}} &
        \multicolumn{2}{c}{\textbf{BART+\sys}} \\
        &
        \multicolumn{2}{p{0.4\linewidth}}{University staff in England are to stage a one-day strike on Wednesday in a row over pay, in a dispute that could lead to further industrial action.} &
        \multicolumn{2}{p{0.4\linewidth}}{Universities have warned of "minor disruption" for students in England after staff rejected an improved pay offer in a dispute over a pay rise.} \\
        \midrule
        \textbf{Ratings} &
        \multicolumn{2}{c}{Consistency: 3 \ Fluency: 3 \ Relevance: 5} &
        \multicolumn{2}{c}{Consistency: 4 \ Fluency: 5 \ Relevance: 3} \\
        \midrule
        \textbf{Annotator}\newline\textbf{Comment} &
        \multicolumn{2}{p{0.4\linewidth}}{\emph{consistency: location abstracted; "one day" on Wednesday not true
        \newline
        fluency: "in a" twice in same sentence, doesn't flow}} &
        \multicolumn{2}{p{0.4\linewidth}}{\emph{consistency: quoted text is not an actual quote; location abstracted
        \newline
        relevance: "improved pay" is misleading, missing key information: "1.1\% marginal improvement"}} \\
        \bottomrule
        \end{tabular}
    \end{threeparttable}
    }
\end{table*}

\end{document}